\documentclass{ieeetran}
\usepackage{cite}
\usepackage{amsmath,amssymb,amsfonts}
\usepackage{algorithmic}
\usepackage{graphicx}
\usepackage{textcomp}
\usepackage{multirow, makecell}
\def\BibTeX{{\rm B\kern-.05em{\sc i\kern-.025em b}\kern-.08em
    T\kern-.1667em\lower.7ex\hbox{E}\kern-.125emX}}

\usepackage{tikz}
\usepackage{textcomp}
\usepackage{hyperref}
\usepackage{lipsum}

\newcommand\copyrighttext{%
  \footnotesize \textcopyright 2024 IEEE. Personal use of this material is permitted.
  Permission from IEEE must be obtained for all other uses, in any current or future
  media, including reprinting/republishing this material for advertising or promotional
  purposes, creating new collective works, for resale or redistribution to servers or
  lists, or reuse of any copyrighted component of this work in other works.
  DOI: \href{<https://ieeexplore.ieee.org/document/10697132>}{10.1109/ACCESS.2024.3469197}}
\newcommand\copyrightnotice{%
\begin{tikzpicture}[remember picture,overlay]
\node[anchor=south,yshift=5pt] at (current page.south) {\fbox{\parbox{\dimexpr\textwidth-\fboxsep-\fboxrule\relax}{\copyrighttext}}};
\end{tikzpicture}%
}

\begin{document}

\title{Body and Head Orientation Estimation from Low-Resolution Point Clouds in Surveillance Settings
\thanks{This work was supported by the National Science Foundation under grant DUE-1928604.

O.N. Tepencelik, P.C. Cosman and S. Dey are with the Department of Electrical and Computer Engineering, and W. Wei was with the Department of Electrical and Computer Engineering, UC San Diego, La Jolla, CA 92093, USA. (e-mail: \{otepence, w8wei, pcosman, dey\} @ucsd.edu)}
}

\author{\IEEEauthorblockN{Onur N. Tepencelik, Wenchuan Wei, Pamela C. Cosman, \textit{Fellow}, IEEE, Sujit Dey, \textit{Fellow}, IEEE}}

\maketitle
\copyrightnotice

\begin{abstract}
We propose a system that estimates people's body and head orientations using low-resolution point cloud data from two LiDAR sensors. Our models make accurate estimations in real-world conversation settings where subjects move naturally with varying head and body poses, while seated around a table. The body orientation estimation model uses ellipse fitting while the head orientation estimation model combines geometric feature extraction with an ensemble of neural network regressors. Our models achieve a mean absolute estimation error of 5.2 degrees for body orientation and 13.7 degrees for head orientation. Compared to other body/head orientation estimation systems that use RGB cameras, our proposed system uses LiDAR sensors to preserve user privacy, while achieving comparable accuracy. Unlike other body/head orientation estimation systems, our sensors do not require a specified close-range placement in front of the subject, enabling estimation from a surveillance viewpoint which produces low-resolution data. This work is the first to attempt head orientation estimation using point clouds in a low-resolution surveillance setting. We compare our model to two state-of-the-art head orientation estimation models that are designed for high-resolution point clouds, which yield higher estimation errors on our low-resolution dataset. We also present an application of head orientation estimation by quantifying behavioral differences between neurotypical and autistic individuals in triadic (three-way) conversations. Significance tests show that autistic individuals display significantly different behavior compared to neurotypical individuals in distributing attention between conversational parties, suggesting that the approach could be a component of a behavioral analysis or coaching system.
\end{abstract}


\section{Introduction}

\IEEEPARstart{B}{ody} and head orientation estimation are fundamental challenges in computer vision, mainly investigated in the context of pedestrian protection and movement prediction \cite{rehder2014head}, along with applications in robotics \cite{lewandowski2019deep} and behavior analysis \cite{chen2012we}. Most work on body and head orientation estimation uses RGB cameras for their low cost and prevalence \cite{chen2012we,kohari2018cnn}, but more expensive RGB-D cameras such as Microsoft Kinect and Intel RealSense have also been used \cite{fanelli2011realdepth, borghi2018face}. However, use of RGB cameras raises privacy concerns in many cases. Studies suggest that people's concerns over privacy have been increasing, with privacy protection mechanisms getting more attention \cite{stark2020don, bhave2020privacy}. We propose a system that uses point cloud data from LiDAR sensors to estimate body and head orientations while protecting user privacy. While depth maps also preserve privacy, most common depth sensors are RGB-D that record color information as well, whereas LiDAR solely outputs depth, making it a more privacy-safe device \cite{joo2022indoor}. There has been increased adoption of LiDAR sensors with declining costs \cite{skyquest}. Many recent projects in different fields such as healthcare \cite{bouazizi2021activity, joo2022indoor}, security, and surveillance \cite{ gunter2020privacy, rodrigues2021laflector}, have adopted LiDAR sensors over privacy invading alternatives. With recent advances in LiDAR technology and big data management systems that enable data scalability \cite{lokugam2022scalability}, they are likely to become more prevalent in stores, workplaces, and hospitals.

\begin{figure*}[!h]
  \centering
  \begin{minipage}[b]{\textwidth}
    \includegraphics[width=\textwidth]{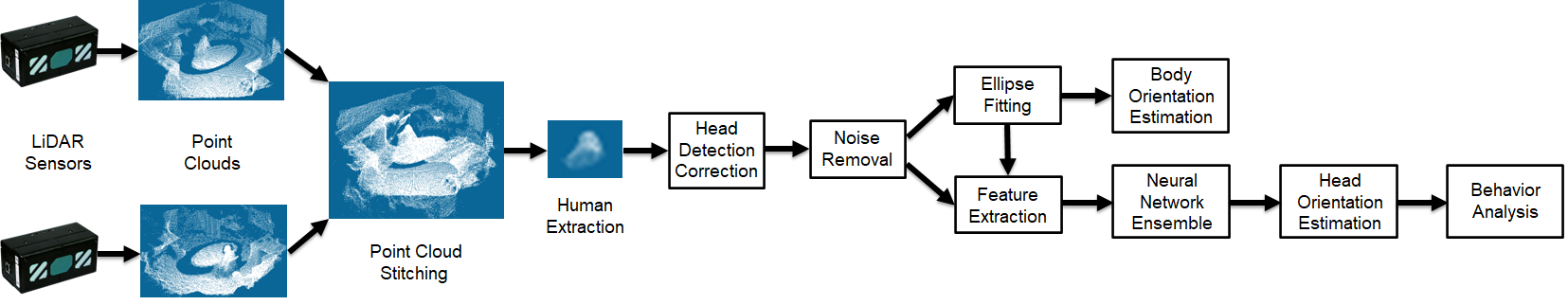}
    \caption{System overview.}
    \label{overview}
  \end{minipage}
\end{figure*}

Available depth image-based models using RGB-D sensors or LiDAR seemed to be good candidates for our need to estimate body and head orientation. However, these models require the sensor to be placed in front of the person, with specific optimal ranges for distance and height, which we refer to as a {\it frontal setting}. In contrast, our system does not require the subject to appear head-on in front of the sensor. Our sensors are placed near the ceiling, looking down at about 45 degrees, and the subject can have arbitrary orientation in the conference area; we refer to this as a {\it surveillance setting}. Our models for body and head orientation estimation with LiDAR sensors are the first that permit a surveillance viewpoint. 

In general, surveillance settings produce low-resolution data; a subject farther from the sensor is represented with fewer points in a point cloud or fewer pixels in an RGB image. Especially for head pose estimation, most models \cite{papazov2015real, padeleris2012head} use high-resolution 3D scans of the head, taken by a sensor close to the subject. With such a setting, it is possible to capture small facial geometric details of the nose tip, eye holes, and chin, which can play a huge role for orientation estimation. While those models are successful for high-resolution data, they face challenges in our case, as our sensors are unobtrusively distant from the people, and it is difficult to identify small facial geometry features due to the decreasing resolution and increasing noise with distance. 

Being able to work with low-resolution data is essential for models targeting a surveillance setting. For example, pedestrian protection applications that aim to detect pedestrians and predict their movements from a surveillance viewpoint or from a sensor mounted on a vehicle would benefit from a system that enables estimations from low-resolution data \cite{rehder2014head, wang2022lidar}. Similarly, Chen \textit{et al.} \cite{chen2012we} proposed a head and body orientation estimation model to analyze and predict behaviors in public spaces such as airports, which could be useful for public safety. Various other studies proposed leveraging head orientation estimation systems for attention and interaction modeling, for instance in museums to analyze which artworks are getting more interest \cite{brvsvcic2013person} or in shopping centers to gauge which products are attracting more customers \cite{liu2007customers}, or in work environments to analyze social interactions \cite{chen2011discovering}.

In this paper, we also present an application of our head and body orientation estimation models. Multiple studies \cite{ba2006study,stiefelhagen2001estimating} have shown that head orientation is a good indicator of visual focus of attention, without the need to estimate gaze orientation. Body and head orientation and movement provide important means of nonverbal communication for fluent social interaction. Individuals with social communication differences (for example, some individuals with Autism Spectrum Disorder (ASD)) might not regularly provide normative nonverbal communication cues, such as periodically making eye contact with a speaker and maintaining a body orientation generally towards them \cite{bal2019autism}. Differences from society's workplace communication norms are one reason that high-functioning young adults with ASD have high unemployment rates \cite{chen2015trends} despite often holding college degrees, average to high IQs, and various useful skills. Furthermore, it was found that many autistic people were terminated from jobs due to communication differences \cite{muller2003meeting}. 

To analyze behaviors related to body and head orientation, we use a triadic (three-way) conversation setting with two \textit{interviewers} and one \textit{subject} sitting around an oval conference table. Triadic conversations are common in professional and social settings and they are harder to navigate compared to dyadic (two-way) interactions \cite{greene2013beyond}. While some autistic individuals may find it challenging to show attention in a dyadic interaction with potential distractions such as objects, a triadic interaction involves an additional person who is a part of the conversation and may require attention. Adjusting body and head orientation in triadic settings is important to engage with both of the other people and make everyone feel included in the conversation \cite{nagels2015feeling}. 

In triadic interactions, some individuals with ASD tend to fixate on one person while ignoring the other for some time \cite{mcparland2021investigating}, a non-normative distribution of attention which could be seen as non-inclusive or socially inappropriate. Other neurodivergent behaviors commonly displayed by autistic individuals include not making eye contact with any of the interviewers while speaking, or not paying attention to a speaker while listening \cite{american2013diagnostic,world1992icd}. Our body and head orientation estimation system can quantify such behavioral differences between autistic and neurotypical individuals. We plan to extend our system to provide coaching and feedback to autistic individuals, imitating the coaching advice of a professional behavioral coach, with the motivation of supporting autistic individuals in practicing conversational engagement skills in preparation for job interviews and workplace communications \cite{tepencelik2021body, artiran2021hmm}. 

Our main contribution in this paper is the development of novel body and head orientation estimation models specifically designed to work with low-resolution point cloud data, generated by two indoor LiDAR sensors from a surveillance viewpoint. Fig.~\ref{overview} shows the system overview. A preliminary version of this work \cite{tepencelik2021body} estimated head yaw angle using a limited set of body and head poses involving a motionless subject with a straight body, lowered arms, and a head pose with no roll or pitch rotations. The current work, enhancing the model in \cite{tepencelik2021body}, is able to estimate orientations while the subject moves naturally and displays various body and head poses. The enhanced model can estimate yaw with similar accuracy even in the presence of roll and pitch rotations. To the best of our knowledge, our head orientation estimation model is the first to estimate orientations from a surveillance viewpoint (low resolution), using LiDAR sensors. Our second contribution in this work is quantifying differences in orienting behavior between neurotypical and autistic individuals using an automated system. Although some of these differences are generally known to characterize ASD, we are the first to quantify them using an automated system, and the first to quantify them in a triadic conversation setting.

The rest of this paper is organized as follows. Section \ref{related} presents an overview of existing literature on body orientation estimation, head orientation estimation and orientation behavior analysis of autistic individuals. Section \ref{dataset} explains our data collection, labeling and cleaning procedures. Section \ref{method} presents our methodology starting from data pre-processing and correction, detailing our body orientation estimation procedure as well as our feature extraction and machine learning approach for head orientation estimation. Section \ref{results} presents the performance of our models in terms of MAE and the comparison with the state-of-the-art. Section \ref{results} also presents an application of our models where we show significant differences between neurotypical and autistic individuals in terms of orientation and attention distribution behavior. Section \ref{conclusion} concludes with a discussion of the current work and our future directions.

\section{Related Work} \label{related}

Body and head orientation estimation are well studied tasks in  computer vision. In this section, we categorize related work according to the data types (RGB images or depth maps/point clouds) as well as the experiment setting (frontal or surveillance). We then introduce a few studies that analyze and compare orienting behaviors of neurotypical and autistic individuals in similar experimental settings.

\subsection{Body Orientation Estimation}

Among the many RGB image-based models for body orientation estimation which are generally in the context of smart vehicles and robotics for human-robot interactions, there are a few which match our type of surveillance setting. Chen \textit{et al.} \cite{chen2012we} proposed a semi-supervised model on RGB images to analyze behavior and attention based on estimated body and head orientations of people waiting for luggage in an airport. The authors of \cite{chen2012human} and \cite{liem2013person} proposed template matching models that combine 2D images from multiple surveillance viewpoints to make 3D orientation estimates. Studies targeting pedestrian orientation \cite{liu2015online, raza2018appearance} usually approach the problem as a classification task, providing less precision compared to regression models.  Many studies such as \cite{flohr2015probabilistic, ardiyanto2014partial} incorporated motion information and tracking techniques into their models as they approach the task from the perspective of a vehicle.  The authors of \cite{weinrich2012estimation, kohari2018cnn} proposed models to estimate body orientation for human-robot interactions, which resemble a frontal setting.

The authors of \cite{brvsvcic2013person} proposed a person-tracking system with a body orientation estimation feature using depth sensors from surveillance viewpoints. The authors use Principal Component Analysis (PCA) on projected point clouds to estimate body orientation. Other than \cite{brvsvcic2013person}, the works on body orientation estimation using depth sensors do not use surveillance scenarios. Shimizu \textit{et al.} \cite{shimizu2016lidar} proposed a model which combines shape and motion information using a LiDAR-mounted robot. Similarly,  \cite{dewantara2022estimating} combines Histogram of Oriented Gradient (HOG) features with motion information tracked by a Kalman filter, using depth images from a Kinect sensor. Other studies use depth along with color information; \cite{liu2013accurate}  enhanced features extracted from an RGB image using depth and motion information, while \cite{ji2017fast} combined features extracted from both RGB and depth images. Experimenting with different CNN architectures that use RGB input, depth input and RGB-D input, authors of \cite{lewandowski2019deep} argued that depth maps are more suitable for estimating orientation than RGB images.

\subsection{Head Orientation Estimation}

Many papers estimate all three Euler angles (yaw, pitch, roll) to define a full head pose \cite{fanelli2011realdepth, borghi2018face}. Depending on the application, such as behavior prediction on pedestrians \cite{chen2012we, raza2018appearance}, pitch and roll angles are often neglected as yaw angle defines the direction people are looking. For our application on the division of attention between two other people in a triadic conversation, we likewise focus on yaw angle. 

As with body orientation, the majority of models in the literature use RGB images, but some use depth, and only a few consider the task in a surveillance setting. Before deep learning techniques, good results were achieved by \cite{fu2006graph, balasubramanian2007biased} using graph embedding, manifold learning and locally linear embedding techniques. Zhao \textit{et al.} \cite{zhao2002real} used a neural network followed by more complex architectures such as random regression forests \cite{fanelli2011real}, deep neural networks \cite{ahn2014real, ahn2018real}, convolutional neural networks (CNN) \cite{hsu2018quatnet, hu2021deep, liu2021anisotropic} and Graph-CNNs \cite{xin2021eva}. The authors of \cite{abate2020head, barra2020web} proposed a technique called web-shaped model to estimate head orientation using 68 facial landmarks which are detected using \cite{kazemi2014one}. A recent study by Yao \textit{et al.} \cite{yao2024head} showed that state-of-the-art performance can be achieved by using only seven of those 68 landmarks, four of which are the corners of the eyes. All these papers target a frontal setting. The authors of \cite{chen2012we, zhang2006head} created various models and surveillance settings for the head orientation estimation task using RGB cameras, as the latter took advantage of extensive research on face detection in a room surveyed by four cameras creating multi-view representations. The authors of \cite{rehder2014head} proposed RGB image based head orientation estimation models, to ensure pedestrian safety from the perspective of a vehicle. Similarly, a CNN-based model in \cite{raza2018appearance} estimated pedestrian orientations from a surveillance viewpoint. While proposing a transfer learning approach, the authors of \cite{k2012adaptation} published the DPOSE dataset, a dynamic, multi-view head pose dataset collected in a room with 4 cameras at surveillance viewpoints. Various papers \cite{hong2018multimodal, Yan_2013_ICCV} proposed new methods and published results on the DPOSE dataset. A CNN-based model in \cite{berral2021realheponet} works on unconstrained RGB images, similar to a surveillance viewpoint with higher resolution. 

Head orientation estimation using depth cameras has a longer history 
(see \cite{malassiotis2005robust}) compared to body orientation estimation using depth cameras. Following the availability of consumer-level depth cameras such as Microsoft Kinect and Intel RealSense, various head orientation estimation models using depth images \cite{kondori20113d, fanelli2011realdepth, padeleris2012head, papazov2015real, ghiass2015highly, martin2014real, borghi2018face} were proposed. The BIWI benchmark \cite{fanelli2011realdepth} contains around 15000 samples of head poses recorded with a sensor placed frontally about 1 meter from the subjects, and it was used for many results \cite{padeleris2012head, papazov2015real, borghi2018face, hong2018multimodal, xu2022head, xin2021eva}. A particle swarm optimization approach was used in \cite{padeleris2012head}, while \cite{papazov2015real} used triangular surface patches as hand-crafted 3D features to estimate orientation. More recent papers such as \cite{borghi2018face, xu2022head} proposed CNN architectures to tackle the problem; \cite{xu2022head} resembles our work as their model uses 3D point clouds as input as opposed to the more common 2D depth maps. Point clouds were also used in \cite{hu2020robust} which leveraged the PointNet++ architecture \cite{qi2017pointnet++} by using its abstraction layers as a feature extractor, and they further improved their work by including temporal information using an LSTM network in \cite{hu2021temporal}. Similar to the depth information based body orientation estimation models, the models listed above assume that the depth sensor is directly in front of the person.

To our knowledge, we are the first to estimate head orientations using a depth sensor from a surveillance viewpoint. Existing models use either RGB images with a surveillance setting, or use depth images with a frontal setting. In this study, we propose models for both body and head orientation estimation.

\subsection{Analysis of Orientation Behavior}

A core diagnostic feature of ASD is differences in social attention \cite{chita2016social}, which include social orienting, joint attention, eye contact, and non-verbal gestures. In this section of our literature review, we mainly focus on orientation behavior, specifically in triadic settings.

As suggested by \cite{clifford2009dyadic}, early triadic behaviors are important for the development of later social responsiveness. The authors of \cite{mcparland2021investigating} studied triadic conversations with low communicative intent (researchers speaking primarily with each other, with occasional input from a child) and dyadic conversations with high communicative intent (a researcher directly interacting with a child) and found that children with ASD made 57\% more gaze fixations to people's faces in these triadic conversations compared to the dyadic ones; the reverse pattern was found for typically developing (TD) children. The authors also found that children with ASD spent 12.3\% less time looking at other people's faces in these triadic conversations compared to the dyadic ones, and 9.7\% less compared to TD children.

Other studies such as \cite{dawson2004early,jarrold2013social} with different experimental settings also provide insight into orienting behaviors of people with ASD. The authors of \cite{dawson2004early} found that children with autism were significantly less likely to respond to social stimuli (such as calling the child's name, or snapping fingers) with a re-orientation of the head, compared to their responses to non-social stimuli (such as a phone ringing), as well as compared to the responses of TD children. In a virtual public speaking experiment, the authors of \cite{jarrold2013social} found that high-functioning children with ASD made contact with the listeners less frequently compared to TD children.

A model to analyze head movement features such as rotation range and frequency in autistic children during face-to-face interactions was proposed in \cite{zhao2021atypical}. The authors reported that compared to TD children, autistic children had a significantly higher level of head movement stereotypy (repetitive, ritualistic head movements), as well as higher rotation range and frequency. Based on these results, the authors developed a machine learning model to diagnose autism in children using the proposed head movement features \cite{zhao2021identifying}.

Many researchers have studied social modulation of gaze, which is the change in gaze orientation based on conversational role (e.g., speaker or listener). In dyadic conversations, listeners generally gaze more at speakers compared to speakers looking at listeners\cite{argyle1976gaze, vertegaal2001eye}. However, \cite{vertegaal2001eye} found that in group conversations, the gaze levels of speakers come close to that of listeners. The authors argued that one reason for this change was that speakers, when addressing a group, need to collect visual feedback from each individual and to maintain the signal that they are addressing each individual.

\section{Data sets} \label{dataset}

Our system uses two ToFv2 LiDAR sensors from Hitachi Vantara \cite{hitachi}. The sensors capture depth information and create a point cloud based on the Time-of-Flight principle \cite{lange2001solid}. We placed sensors at opposite ceiling corners in a 3x3.5 meter conference room, looking down on an oval table. The point clouds are stitched together using rotation and translation. For both the static and conversation datasets, we ensured with calibration tests that the sensor positions, orientation angles and stitching parameters are the same before each data collection session for data reliability. For the static dataset, the sensors were manually calibrated by visual inspection of the output point clouds. We refined the calibration procedure for the conversation dataset using the fixed locations of four pieces of reflective tape that provide stronger sensor signal. We use the sensor software's built-in human detector which outputs XY-coordinates of the center of gravity of the detected human, as well as a Z-coordinate of the top of the head, Z$_{head}$. In this section, we present our data collection, labeling and cleaning procedures. This study was approved by the UC San Diego Institutional Review Board (Protocol 210775, Date 7/1/2021).  

\subsection{Static Dataset}

In \cite{tepencelik2021body}, we created a static dataset from 15 neurotypical adults with and without glasses and face masks, and with varying hairstyles and heights. Our dataset consisted of 8 male and 7 female subjects with an average age of 26.2. We collected data with one subject at a time, while the subject follows guidance arrows (as ground truth) placed on a table. Each subject orients their head towards 13 predetermined angles (-90 to +90 degrees, in increments of 15 degrees). For capturing each point cloud, the subjects are instructed to be motionless with their upper body straight to the front and parallel to the table edge, and with their hands on the table or on their lap. Point clouds corresponding to each head orientation angle were captured one by one as snapshots, rather than getting sampled from a continuous data stream. 

\begin{figure*}[!h]
  \centering
  \begin{minipage}[b]{\textwidth}
    \centering
    \includegraphics[width=\textwidth]{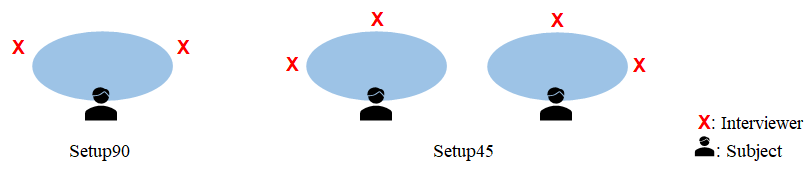}
    \caption{Conversation Setups}
    \label{setups}
  \end{minipage}
\end{figure*} 

\subsection{Conversation Dataset}

While the static dataset was useful in the early stages of model development, we found that a model trained on it struggled to accurately estimate head and body orientations of people engaged in real-world conversations, which involve natural movement and varying head and body poses. To create a dataset that represents natural aspects of a conversation, we recorded conversations in a triangular conversational setting with two interviewers and one subject. The subjects were 12 neurodivergent individuals who had received a community diagnosis of ASD and 8 neurotypical individuals. The 12 autistic subjects consisted of 10 males and 2 females with an average age of 22.1 while the 8 neurotypical subjects consisted of 6 males and 2 females with an average age of 23.6. Each subject participated in 2 sessions of 8 to 15 minutes in two different seating setups as shown in Fig.~\ref{setups}. In $Setup90$, the interviewers are separated by an angle of around 90 degrees (ranging from 75 to 105 across sessions) from the subject's perspective, whereas in $Setup45$, the separation is around 45 degrees (ranging between 35 and 55 degrees). Different seating positions helped us collect data with different head orientations, creating useful variety in the dataset.

The subjects were asked to engage naturally as they would in a casual conversation, and they were not informed prior to the session that the data would be used to analyze head and body orientation. The conversation starts with a casual question such as “What do you do in your free time?” and continues based on the answers of the subject. It is intended that the subject is the main speaker throughout the conversation, while the interviewers listen in an engaged way, while also making brief comments, asking follow-up questions, and shifting topics. The duration, pace, topics, and conversational roles were controlled to the best extent possible to prevent these external factors from confounding the behavior analysis portion of the study. The interviewers followed a conversation script reasonably closely so that each participant was asked questions about the same set of topics, in the same order. During the sessions, LiDAR point cloud data were recorded with an average frame rate of 1.5 fps. We also recorded RGB video solely for ground truth labeling and not for model development. Unlike the static dataset, in the conversational setting we observed many different body poses by the subjects, such as turning their upper body towards one interviewer, using their arms and hands as part of their body language, and putting their hands close to their face while thinking or listening.

\subsection{Data Extraction and Labeling}

To create a dataset that contains various head poses that occur during a real conversation, we manually sampled and labeled data from the sessions. To manually estimate the ground truth head orientation from video snapshots, we used 3 reference orientations. Two of these are computed using the point cloud centroid coordinates of the two interviewers with respect to the subject, at the time of the snapshot. The third reference angle is the average of the first two, representing the midpoint of the two interviewers. For example, if one interviewer is seated 30 degrees to the left of the subject, and the other interviewer is seated 50 degrees to the right, the three reference angles are +30, -50 and -10 degrees. If the subject's head is oriented directly towards an interviewer, the ground truth label is the reference angle for that interviewer. If the subject is looking at the midpoint of the two interviewers (a common situation when speaking to multiple listeners), the ground truth label is the midpoint reference angle. The manual sampling and labeling procedure is detailed below:
\begin{enumerate}
  \item Align the RGB video and point cloud recordings based on timestamps.
  \item From the video recording, identify an instance where the subject’s head orientation is static for at least two seconds and close to a reference point.
  \item Extract the point cloud data that corresponds to the identified video instance.
  \item Calculate the reference angles using the point cloud coordinates of the subject and the interviewers.
  \item Estimate the subject's head orientation from the video as ground truth, with the help of reference angles.
\end{enumerate}
From this, we obtained 80 to 140 instances from each ASD subject, totaling 1400 point cloud frames. We ensured a variety of body and head poses in the dataset, including challenging ones such as subjects with their hands on their face or chin, arms over their head or their bodies heavily leaning towards the table, the back or the sides. We also tried to ensure that the frequencies of various different poses within the dataset are reasonably close to how often each pose is displayed by the subjects. We achieved this by sampling a data point every time a subject shifts their pose (e.g. turns from one interviewer to the other, leans towards the table, raises their arm, places their hands on their face etc.), and sampling data points periodically (one sample every 10 to 15 seconds) if the subject's pose is stable for longer periods. Note that sampling a data point is only possible if there is an instance where the subject's head orientation is static for at least two seconds and close to a reference point, as stated by the second item in the above data sampling procedure. It is also important to note that this process is done only once to establish ground truth for creating the model, and should not be construed as a calibration step needed in subsequent use of the model.

This manual sampling and labeling is subject to potential human errors. Estimating head orientation accurately from a video is hard, although we sampled instances where the subject’s head is oriented towards an interviewer or the midpoint, which are relatively easier positions to interpret the orientation.  To quantify the human labeling error, we conducted a simple experiment on three subjects. Using guidance arrows to guide the subject to adjust their head orientation in 5 degree increments, we collected a random sequence of head orientations. On average, human labels based on video recordings differed by about 4.5 degrees from the guidance arrow ground truth. A portion of this error comes from the subjects imperfectly aligning their heads with the guidance arrows, which is a potential issue that exists in the static dataset as well.

\subsection{Data Cleaning}\label{cleaning}

Our dataset presented challenges in data cleaning. First, there were issues caused by the sensors. Network overload during real-time data collection caused lost point cloud frames that were replaced by previously recorded frames, an issue present in 3 of 24 data collection sessions. Of the 1400 point clouds manually sampled from the real-time sessions, 56 were excluded as they were sampled from a repetition sequence and therefore did not reflect the true state of the environment at the matched timestamp. Another LiDAR sensor issue was the inaccuracy of the built-in human detection algorithm. The sensor may confuse the subject’s shoulder with their head, resulting in a wrong output of center of gravity and $Z_{head}$. In that case, the wrong portion of the human body is cropped out and the point cloud lacks points from the other shoulder. We removed an additional 74 point clouds due to this erroneous human detection. In Section \ref{headpos}, our proposed head position estimation algorithm can mitigate this issue. We also used our proposed algorithm to detect the instances where this issue happened and eliminate them if the discrepancy between the built-in head position and our estimated head position is bigger than 10 cm either in the horizontal plane or in the vertical axis. 

Secondly, there are human errors during manual sampling and labeling, such as selecting a wrong timestamp from the video recording, or a slight lack of synchronization between the video recording and point cloud sequence, leading to the selection of the wrong point cloud frame. In such instances, we observed that the next or previous point cloud is more suitable for the suggested head orientation label, indicating that the wrong point cloud was sampled and the subject's head orientation changed within consecutive frames. We address this problem by using our model's predictions on the neighboring frames as a preliminary indication of a wrong sampling or a synchronization issue, and replace the point cloud with its neighbor if we can visually confirm the issue from the video recording and point cloud sequence. In Fig.~\ref{ellipse}, two consecutive point clouds from one of our data sequences are shown. The human labeler originally sampled Fig. \ref{ellipse}{a} from the sequence, trying to match with the video instance where the subject's head was towards an interviewer seated 35 degrees to the subject's left. However, due to a slight synchronization issue, the point cloud in Figure \ref{ellipse}{a} belongs to the middle of the head movement towards that interviewer, and the next point cloud (shown in Figure \ref{ellipse}{b}) should have been sampled instead. The head movement becomes apparent when 4-5 consecutive point clouds are visualized on top of each other and compared with the corresponding video sequence, which allows one to choose the point cloud matching the intended orientation label. Of the remaining 1270 point clouds, 41 were replaced by their neighboring point clouds due to this labeling issue.

\section{Methodology}\label{method}

In this section, we present our data pre-processing steps and body and head orientation estimation models. Section \ref{preproc} details the modified noise removal algorithm from \cite{tepencelik2021body} and Section \ref{headpos} presents a head position correction procedure which led to improvements in model performance. Section \ref{boe} presents our body orientation algorithm. Section \ref{features} describes our hand-crafted geometric features and feature elimination procedure while Section \ref{hoe} presents our head orientation estimation pipeline.

\subsection{Pre-Processing}\label{preproc}

For this work, we introduced additional pre-processing steps compared to our work in \cite{tepencelik2021body}. As explained in Section \ref{headpos}, the built-in head height estimation is often inaccurate, so we propose an improved estimation procedure to obtain the head height, $Z_{head}$. To extract the region of interest which consists of the upper body and head, we crop a cylinder-shaped boundary around each person’s point cloud using the centroid and a radius of 50 cm. For each subject, a threshold for the upper body set empirically as the top 27\% of their height (in a seated position) is computed from our improved estimated $Z_{head}$. We also removed points from the head point cloud if they are at least 15 cm away from the head center and from the upper body point cloud if they are at least 25 cm away from the body center on the horizontal plane, after separating the head and upper body point clouds. The computation of head and body centroids and the separation of head and upper body point clouds are explained in Section \ref{method}. With these additional steps, we were able to remove points that did not belong to the region of interest and instead belong to the table, the back of the chair, or noise, which created many distortions in our preliminary work \cite{tepencelik2021body}.  

The upper body point clouds obtained from the pre-processing step consist of about 1800 points on average, varying between about 1500 and 2100 points per case. Since our system is in a surveillance setting, our upper body point clouds have lower resolution compared to other work, e.g., the BIWI dataset \cite{fanelli2011realdepth} contains around 10,000 points for a person's face alone. Estimating body and head orientation from low-resolution LiDAR data is challenging due to the lack of detail in the small region of interest. Moreover, the point cloud data from the surveillance angle are noisy, especially from hair and other complex features on the head. To mitigate this, we apply a k-nearest neighbor noise removal step, where we delete a point if the average distance between the point and its 10 nearest neighbors is larger than 50 mm. All the parameters and thresholds presented in this section are treated as hyperparameters which were optimized during the training of our head orientation estimation model. We initialized each parameter based on our visual and statistical analysis of the data, and optimized them for model performance.

\subsection{Head Position Correction}\label{headpos}

Although the LiDAR's built-in human detection capability usefully extracts human point clouds from the environment point cloud, it does not pinpoint the head center in the horizontal plane as the center of gravity of the human is not necessarily the same as their head center. The algorithm also provides inconsistent results for $Z_{head}$. The two sensors make independent estimates which are averaged to form a joint estimate, which is usually better than relying on a single sensor estimate. However if one sensor makes a large estimation error, the joint estimation is not good enough to recover. Accurate and consistent estimation of $Z_{head}$ across the whole dataset is especially important as it is used to separate the head and upper body point clouds.

We improved the estimate of the head center and $Z_{head}$ from the point cloud. If $Z_1$ and $Z_2$ (in centimeters) represent the built-in $Z_{head}$ estimates for sensors 1 and 2, we use $Z = ((Z_1 + Z_2) / 2) - 15$ as the initial separation threshold; points above this threshold belong to the head and points below belong to the upper body. This yields two disjoint point clouds, $PC_{head}$ and $PC_{body}$. We project the points in $PC_{head}$ onto the horizontal plane and use least-squares ellipse fitting on them, as detailed in the following section. The ellipse center is a more accurate estimate of the head center, compared to the built-in estimate from the LiDAR sensors. $Z_{head}$ is determined by sorting the points by their z-coordinate and taking the highest point with a maximum height difference of 1mm with the next 5 highest points. This operation mitigates noise distorting the $Z_{head}$ calculation, and generally pinpoints the top of the head where the height difference between points should be saturated. The head center computed from the least-squares ellipse, together with this $Z_{head}$, represent the center point of the subject’s top of the head.

This improves our preliminary work \cite{tepencelik2021body}, which relied heavily on the built-in estimates. In  \cite{tepencelik2021body}, the inaccurate built-in estimation for $Z_{head}$ was used to base the separation threshold to obtain $PC_{head}$ and $PC_{body}$, which sometimes caused $PC_{head}$ to contain points from the shoulders or $PC_{body}$ to contain points from the chin.

\subsection{Body Orientation Estimation} \label{boe}

The body orientation estimation model is a geometric model which takes advantage of the ability to change the viewpoint from which a point cloud is seen, and uses the birds-eye view of the room. The cropped point clouds are projected onto the horizontal plane. After estimating $Z_{head}$, we separate head and body points using a refined threshold of $Z = Z_{head} - 17.5$ and calculate the 2D ellipse that best fits the projected $PC_{body}$ based on least squares error, with the long axis of the ellipse representing the frontal (shoulder-to-shoulder) axis. We use the conic representation of an ellipse:
\begin{equation}
    E(x,y) = ax^2 + bxy + cy^2 + dx + ey + f = 0 \label{eq}
\end{equation}
The optimal coefficients are estimated using the direct least squares ellipse fitting method by Fitzgibbon \textit{et al.} \cite{fitzgibbon1999direct}. The noise removal pre-processing is important for this procedure to work well, as noise points that are generally on the edges may result in large squared errors. The correction of the built-in $Z_{head}$ estimation is also crucial as explained in the previous section.

After the frontal axis is determined, there remains the issue of which side of the ellipse is the front. We calculate the average perpendicular distance of each point in $PC_{head}$ from both sides to the frontal axis. Assuming that a person's head is almost always in front of their body (their frontal axis), the front is taken as the side with higher average perpendicular distance. In our datasets, this assumption holds true 99.7\% of the time and the front side is correctly determined.

\begin{figure}[ht]
  \centering
  \begin{minipage}[b]{0.47\textwidth}
    \centering
    \includegraphics[width=\textwidth]{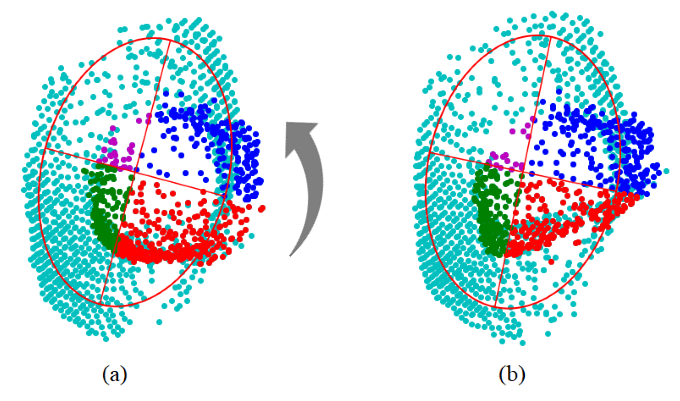}
    \caption{Least squares ellipse fitting for body orientation estimation via the long axis of the fitted ellipse, which also determines the four quadrants of the head relative to the body. Light blue points are projected upper body points (shoulders and chest); dark blue, red, purple and green points are projected head points, representing the four quadrants, in order. The first and second quadrants represent the front-left and front-right sides of the head, respectively; while the third and fourth quadrants represent the back-left and back-right sides of the head. (a) Head orientation label unknown; point cloud belongs to a head movement that starts at 0 degrees and moves to the left (b) Head orientation labeled as 35 degrees to the left.}
    \label{ellipse}
  \end{minipage}
\end{figure}

\subsection{Feature Extraction and Selection} \label{features}

For head orientation estimation, simple geometric approaches were not sufficient as details of facial features are not accurately captured by the sensors. Therefore, we engineered our own geometric features from the point clouds. The surveillance setting allows low-level geometric features but, given the low-resolution nature of our data, higher level 3D features such as surface patches \cite{papazov2015real, fanelli2011realdepth} or curvatures \cite{segundo2010automatic} did not produce useful results on our dataset. The approach proposed in \cite{hu2020robust}, using the initial layers of PointNet++ to extract feature representations, or a Graph-CNN approach proposed in \cite{xu2022head} similarly did not suit our data. However,  as shown in our prior work \cite{tepencelik2021body} and the current study, our low-level geometric features allow estimation of head orientation with reasonable accuracy in a low-resolution surveillance setting. 

The feature extraction is done after noise removal and ellipse fitting to the upper body. The upper body ellipse divides $PC_{head}$ into four quadrants which produce supportive features for the model, based on the point locations with respect to the body center. Fig.~\ref{ellipse} shows two projected human point clouds, the optimal ellipse fit for body orientation estimation, and the resulting four quadrants of the head for each of the clouds.

The features we extract are the $(x,y)$ coordinates of the subject's centroid in the sensor coordinate system, as well as a number of features that use a subject-centric coordinate system.  These features are the principal components and the basic distribution properties of the points in $PC_{head}$ (mean, standard deviation, minimum and maximum coordinates), as well as of the points in its four quadrants separately, the estimated nose coordinates based on the centroid of the 10 furthest projected points from the head center, and the axis lengths and orientations of a separate ellipse fitting procedure on $PC_{head}$. Some of the features we extracted emerged from our initial ideas on how to achieve accurate head orientation estimations. For example, the principal components of the head point cloud and the furthest points from the head centroid corresponding to the nose tip were two ideas to directly estimate head orientations. While none of these worked well on their own, they served well as features to a more complicated model. The features extracted from a single point cloud constitute a feature vector with 103 entries (as x and y dimensions produce distinct features).

For a low-resolution regression task, a feature space with over 100 dimensions presents a higher likelihood of overfitting. To mitigate this, we use the Random Forest Recursive Feature Elimination (RF-RFE) process \cite{granitto2006recursive} which involves repeatedly training a random forest regressor, ranking the features according to their importance, and eliminating the least important feature(s) in each iteration. This approach has been successfully used in many studies \cite{misra2020improving,zvarevashe2018gender,ustebay2018intrusion}. After applying RF-RFE , the optimal feature set had 42-dimensions, with principal components proving to be important features along with some engineered features such as the estimated nose position and the head ellipse parameters. The feature elimination procedure revealed that our two initial ideas involving principal components and nose tip estimation were among the most useful features. Other features in the optimal feature set are a mix of head quadrant principal components and distribution properties of the head point cloud in certain dimensions. Intuitively, horizontal dimensions should hold more importance compared to the vertical dimension since we are estimating the head orientation in the yaw axis. Some head quadrants turned out to be more important than others based on where the sensors are located and their angles in which they view the subjects.

\begin{table*}[!h]
\centering
\caption{Conversation Dataset MAE for Head Orientation Estimation after each Data Cleaning and Processing Step}
\label{maetable}
\begin{tabular}{|cl|cc|}
\hline
\multicolumn{2}{|c|}{\multirow{2}{*}{Process}}                 & \multicolumn{2}{c|}{Mean Absolute Error on Conversation Dataset}                                 \\ \cline{3-4} 
\multicolumn{2}{|c|}{}                                          & \multicolumn{1}{c|}{Model Trained with Conversation Dataset} & Model Trained with Static Dataset \\ \hline
\multicolumn{2}{|c|}{Initial Model (from \cite{tepencelik2021body} but trained as specified in column)}                             & \multicolumn{1}{c|}{19.6}                                    & 35.1                              \\ \hline
\multicolumn{2}{|c|}{Head position correction (Sec. \ref{preproc})}                  & \multicolumn{1}{c|}{18.3}                                    & 32.8                              \\ \hline
\multicolumn{2}{|c|}{Removing repeated point clouds (Sec. \ref{cleaning})}            & \multicolumn{1}{c|}{17.2}                                    & 32.8                              \\ \hline
\multicolumn{2}{|c|}{Removing point clouds with missing points (Sec. \ref{cleaning})} & \multicolumn{1}{c|}{15.9}                                    & 29.5                              \\ \hline
\multicolumn{2}{|c|}{Using neural network ensembles (Sec. \ref{hoe})}            & \multicolumn{1}{c|}{14.2}                                    & 26.4                              \\ \hline
\multicolumn{2}{|c|}{Fixing wrongly synchronized point clouds (Sec. \ref{cleaning})}  & \multicolumn{1}{c|}{13.7}                                    & 26.4                              \\ \hline
\end{tabular}
\end{table*}

\subsection{Head Orientation Estimation}\label{hoe}

For head orientation estimation, we use a pipeline of feature extraction and an ensemble of multi-layer perceptron-based regression networks. To train the head orientation estimation model, we use leave-one-out cross-validation, where the point clouds of each subject are used one time as the test set, and used in training otherwise. Thus each autistic subject has their own model that has never seen that subject before. Depending on the subject, each of the leave-one-out models was trained using 1150 to 1200 point clouds from the dataset and tested on 70 to 120 point clouds. On average, each model was trained with 92\% of our dataset. For neurotypical subjects, we use a model trained with the whole dataset of samples from autistic subjects.

We chose to use a neural network based regression model as a result of experimentation with multiple different approaches and algorithms. As discussed in Section \ref{features}, we initially experimented with other proposed head orientation estimation algorithms from the literature. After concluding that the existing approaches are not suitable for our dataset, we extracted our own features and experimented with multiple different machine learning algorithms such as SVM, decision trees, random forests, gradient boosting and neural network based regression. The latter performed best, with random forest regression a close second. We used random forest regression to measure feature importance as discussed in Section \ref{features}.

Neural networks are typically high variance estimators, as was our preliminary model  \cite{tepencelik2021body}. A dataset of noisy low-resolution point clouds leads to even more variance in predictions. To reduce the estimation variance and improve overall model performance \cite{mendes2012ensemble, ren2016ensemble}, we  enhanced our initial model by deploying an ensemble of neural networks, where each individual network was initialized with different random weights. Often, different initial weights are enough to generate significantly different models \cite{mendes2012ensemble, ren2016ensemble}, to create a diverse ensemble. To create the ensemble, we train 20 separate models and rank them based on their performance on the validation set. Then we use Forward Subset Selection \cite{mendes2012ensemble} to select the models as follows. We start with an initial ensemble of 3 best models, and iteratively add the next best model in the pool to the ensemble until the ensemble performance on the validation set does not improve with the addition of a new model. We ended up with ensembles that contain 3 to 8 models, with the mode and median being 6 models.

\section{Results} \label{results}

In this section, we discuss the performance of our proposed models. Section \ref{metrics} evaluates our models based on mean absolute error (MAE). We present MAE values for our models based on the number of features, the selected feature set and an ablation study of each of the pre-processing steps. We also compare our work to two state-of-the-art head orientation estimation models, as well as other existing literature. Section \ref{behavioranalysis} introduces an application of our estimation models, comparing attention distribution patterns of neurotypical and autistic individuals in triadic conversation settings. We find statistically significant differences between the two groups.

\subsection{Error Metrics} \label{metrics}

\setlength{\tabcolsep}{6pt}
\renewcommand{\arraystretch}{1.25} 

We primarily use MAE to evaluate model performance. For the body orientation estimation model applied to the static dataset, with our proposed improvements, we achieve an MAE of 5.21 degrees compared to the MAE of 8.37 degrees reported in \cite{tepencelik2021body}. The model improved significantly with the head position correction presented in Section \ref{headpos}. Our proposed ellipse fitting method for body orientation estimation outperforms the PCA approach proposed by \cite{brvsvcic2013person}, as the latter produced an MAE of 7.95 on our dataset. We found that the ellipse fitting approach is more robust against noise in the point clouds.

We train and evaluate the head orientation estimation model on data sampled from our new conversation dataset, from 12 autistic subjects. We use a different model for each subject, where data from the other 11 subjects are used for training. On average, our new modeling approach produces an MAE of 13.73 degrees across 12 leave-one-out models for head orientation estimation. When the model from \cite{tepencelik2021body} that was trained with only the static dataset is applied to the 12 subjects on the conversation dataset, the MAE is 26.4 degrees due to the challenges caused by different body poses and natural movements in real-world settings. Our new model based on our conversation dataset outperforms our model in \cite{tepencelik2021body} by about 50\% in a conversational setting. In \cite{tepencelik2021body}, we reported an MAE of 12.69 on our experimental static setting, showing that our new model is able to reach similar levels of accuracy in a conversational setting. A more detailed summary is presented in Table \ref{maetable} which shows the development of our final model as well as a comparison with our initial model. 

\begin{figure}[ht]
  \centering
  \begin{minipage}[b]{0.465\textwidth}
    \centering
    \includegraphics[width=\textwidth]{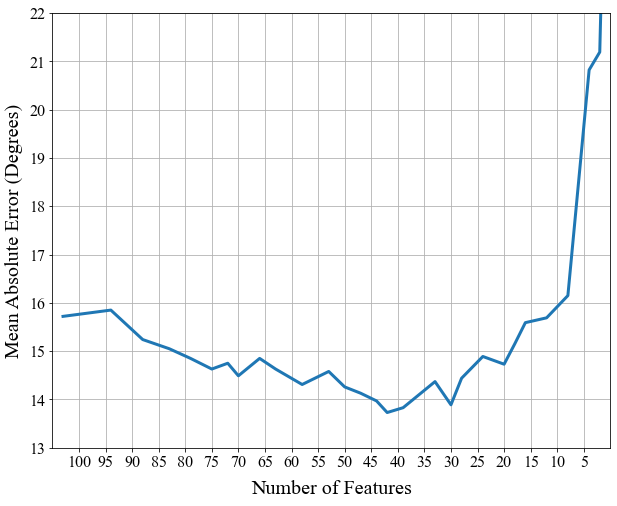}
    \caption{Evolution of mean absolute estimation error with the Random Forest Recursive Feature Elimination procedure. The initial MAE with 103 features is 15.72 degrees, whereas the MAE with only 1 feature left in the feature space is 25.61 degrees. The optimal feature set contains 42 features and leads to an MAE of 13.73 degrees.}
    \label{rfrfe}
  \end{minipage}
  \vspace{-1em}
\end{figure}

\begin{figure}[ht]
  \centering
  \begin{minipage}[b]{0.465\textwidth}
    \centering
    \includegraphics[width=\textwidth]{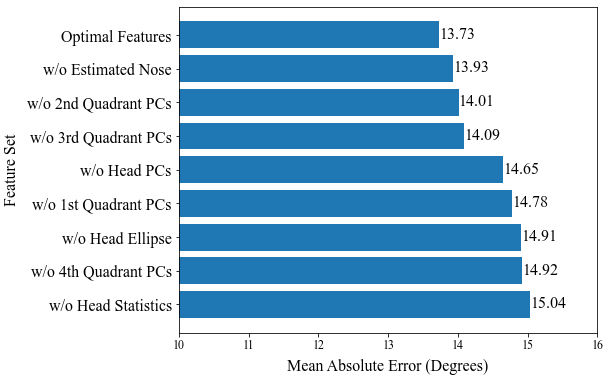}
    \caption{Performance of optimal feature set and its subsets (PC = Principal Component).}
    \label{featureperf}
  \end{minipage}
\end{figure}

\begin{table*}[h]
\centering
\caption{State-of-the-Art Comparisons for Head Orientation Estimation}
\label{behsummary}
\begin{tabular}{|c|c|c|}
\hline
Model & MAE on our dataset & MAE on respective dataset \\ 
\hline
Proposed Neural Network Ensemble & 13.7 & - \\
\hline
PointNet++ Regression \cite{hu2020robust} & 24.2 & 7.32 (MSE) \\
\hline
Graph-CNN PointNet++ \cite{xu2022head} & 23.5 & 1.82 \\
\hline
\hline
HOG-SVM-HMM Pipeline \cite{rehder2014head} & - & 19 \\
\hline
Coupled Adaptive Classifier \cite{chen2012we} & - & 23.6 \\
\hline
Multitask Manifold CNN \cite{hong2018multimodal} & - & 31 \\
\hline
Face Detection-Naive Bayes-HMM \cite{zhang2006head} & - & 33.6 \\
\hline
\end{tabular}
\end{table*}

\begin{table*}[h]
\centering
\caption{Comparison of neurotypical and autistic subjects on two experimental setups}
\label{behaviors}
\begin{tabular}{|cl|cc|cc|}
\hline
\multicolumn{2}{|c|}{\multirow{2}{*}{Statistic}}                          & \multicolumn{2}{c|}{Setup90}         & \multicolumn{2}{c|}{Setup45}         \\ \cline{3-6} 
\multicolumn{2}{|c|}{}                                                    & \multicolumn{1}{c|}{ASD}    & NT     & \multicolumn{1}{c|}{ASD}    & NT     \\ \hline
\multicolumn{2}{|c|}{Average duration of contact}                         & \multicolumn{1}{c|}{11.8}   & 6.3    & \multicolumn{1}{c|}{7.7}    & 6.1    \\ \hline
\multicolumn{2}{|c|}{Maximum duration of contact}                         & \multicolumn{1}{c|}{48.8}   & 16.5   & \multicolumn{1}{c|}{24.6}   & 16.5   \\ \hline
\multicolumn{2}{|c|}{Average duration of NOT contacting anyone}           & \multicolumn{1}{c|}{53.2}   & 41.1   & \multicolumn{1}{c|}{58.6}   & 30.9   \\ \hline
\multicolumn{2}{|c|}{Number of contacts per minute}                       & \multicolumn{1}{c|}{0.84}   & 1.37   & \multicolumn{1}{c|}{1.44}   & 2.15   \\ \hline
\multicolumn{2}{|c|}{Total duration of contact \% during an interview}    & \multicolumn{1}{c|}{11.5\%} & 13.4\% & \multicolumn{1}{c|}{15.6\%} & 16.6\% \\ \hline
\multicolumn{2}{|c|}{Maximum duration of exclusions}                      & \multicolumn{1}{c|}{50.3}   & 16.4   & \multicolumn{1}{c|}{20.3}   & 9.3    \\ \hline
\multicolumn{2}{|c|}{Number of exclusions per minute}                     & \multicolumn{1}{c|}{0.28}   & 0.10   & \multicolumn{1}{c|}{0.19}   & 0.11   \\ \hline
\multicolumn{2}{|c|}{Total duration of exclusions \% during an interview} & \multicolumn{1}{c|}{15.5\%} & 4.2\%  & \multicolumn{1}{c|}{7.4\%}  & 3.1\%  \\ \hline
\end{tabular}
\end{table*}

Fig.~\ref{rfrfe} shows the evolution of model performance (MAE) as we eliminate features with the RF-RFE procedure described in Section \ref{features}. Without RF-RFE, using the whole feature space, the model performance would have been 13\% worse compared to the optimal feature set, with an MAE of 15.72 degrees. The MAE of the model with only 1 feature is 25.61 degrees, 46\% worse than the optimal performance.

After obtaining the optimal feature set using the RF-RFE procedure, we conducted some experiments to analyze feature importance. Fig.~\ref{featureperf} compares model performance in terms of MAE with the optimal feature set and its subsets in which certain features are excluded. The figure shows the importance of engineered features, including the estimated nose position, principal components of the head quadrants, and parameters of the head ellipse.

\subsection{Comparisons with state-of-the-art}

While some papers report smaller errors on head orientation estimation, they use either high-resolution 3D scans of the face when the sensor is placed directly in front of the person \cite{kondori20113d, fanelli2011realdepth, papazov2015real, padeleris2012head, ghiass2015highly, martin2014real, borghi2018face}, or an RGB camera \cite{fanelli2011real, ahn2014real, ahn2018real, hsu2018quatnet}. But for those studies which used indoor RGB surveillance datasets, MAE values of 19, 23.6, 31 and 33.6 degrees were reported \cite{rehder2014head}, \cite{chen2012we}, \cite{hong2018multimodal} and \cite{zhang2006head}, respectively. While we outperform the above studies which used RGB cameras from surveillance viewpoints in terms of raw MAE numbers, it is hard to make exact comparisons as the datasets are not unified and each dataset has its own challenges.

To make fair comparisons, we evaluate the performance of two PointNet++ based state-of-the-art architectures on our dataset. The architecture in \cite{hu2020robust} uses the set abstraction layers of PointNet++ as feature descriptors followed by a fully connected regression layer. The authors of \cite{xu2022head} modified the PointNet++ set abstraction layers with a Graph-CNN approach, proposing a cascaded classification and regression architecture. Both of these architectures rely on PointNet++'s ability of extracting local features from a point cloud. Since these local features may not be easily identifiable in low-resolution point clouds, these approaches did not work as well on our dataset, producing MAE values of 24.2 and 23.5 degrees for \cite{hu2020robust} and \cite{xu2022head}, respectively.

\subsection{Behavior Analysis of Autistic and Neurotypical Individuals} \label{behavioranalysis}

In this section, we quantify some orienting behaviors of autistic and neurotypical individuals in two experimental triadic conversation setups, as shown in Fig.~\ref{setups}.

To quantify behaviors, we define the following two terms: \textit{Contact} and \textit{Exclusion}. Based on studies that suggest ``2 to 3 seconds" \cite{weinschenk2012100, steele2009presentation} or ``a few seconds" \cite{ailes2012you} of eye contact is optimal when addressing multiple people to connect and make them feel included in the conversation, \textit{Contact} in our context is defined as 3 consecutive frames where the head orientation is inside the region of an interviewer, where a frame is about 0.7 seconds and an interviewer's region is defined as $\pm15^o$ from their position.  Making occasional contact with an interviewer is important to make them feel included in the conversation \cite{vertegaal2002explaining} and maintain the signal that they are being addressed \cite{vertegaal2001eye}. \textit{Exclusion} is defined over a 20-frame window; if there are at least 15 estimated head orientations in the region of one interviewer and none in the region of the other, the other interviewer is considered to be excluded from the conversation. In Table~\ref{behaviors}, we present statistics related to \textit{Contact} and \textit{Exclusion} extracted from our conversation sessions using our head orientation estimation model. We present the averages for each statistic across 12 sessions with autistic subjects and across 8 sessions with neurotypical subjects in each setup.

\begin{table*}[h]
\centering
\caption{Summary of statistically significant behavioral differences between the ASD and NT populations}
\label{behsummary}
\begin{tabular}{|c|c|c|c|c|c|c|}
\hline
Behavior & Setup & ASD Mean & NT Mean & t-statistic & p-value & Cohen's d \\ 
\hline
Maximum duration of exclusions & Setup90 & 50.3 & 16.4 & 2.53 & 0.014 & 1.24 \\
\hline
Exclusions \% during an interview & Setup90 & 15.5\% & 4.2\% & 2.69 & 0.011 & 1.41 \\
\hline
Average duration of not making contact with anyone & Setup90 & 53.2 & 41.1 & 1.84 & 0.046 & 0.66 \\
\hline
Number of contacts per minute & Setup90 & 0.84 & 1.37 & -2.21 & 0.023 & 1.08 \\
\hline
Number of contacts per minute & Setup45 & 1.44 & 2.15 & -1.95 & 0.037 & 1.34 \\
\hline
While speaking, looking at the interviewer who spoke last & Setup90 & 30.3\% & 22.2\% & 1.87 & 0.038 & 0.45 \\
\hline
\end{tabular}
\end{table*}

From Table~\ref{behaviors}, we observe that the orienting behaviors of the autistic and neurotypical individuals diverge more in $Setup90$, compared to $Setup45$. When the interviewers are further apart, the autistic individuals have more difficulty with distributing their attention between two conversational partners. Neurotypical subjects tend to make contact with the interviewers in shorter bursts, whereas autistic individuals frequently dwell on one interviewer for a longer period of time. With shorter and more frequent contacts, neurotypical individuals are more likely to ensure that both interviewers feel included in the conversation. Autistic individuals more often have an exclusion, seen through the much higher maximum exclusion duration and total percentage of time spent while an interviewer is excluded.

The authors of \cite{jarrold2013social} suggested that children with ASD made fewer contacts with listeners while speaking to multiple people, compared to typically developing children. Similarly, the authors of \cite{mcparland2021investigating} showed that autistic children spend less time looking at other people's faces in triadic conversations compared to TD children. Our findings are consistent with these, as we observe from Table~\ref{behaviors} that people with ASD made fewer contacts per minute and spent less total time in contact with the listeners. 

We examined statistical significance with independent sample t-tests on the data of the two groups. An independent t-test suggests that there is a significant difference between the averages of two groups if the $p$-value (the probability of this difference occurring by chance) is smaller than the widely accepted threshold of 0.05. We also report Cohen's $d$ values associated with each statistic as the effect size, which is the difference between the group means divided by the pooled standard deviation \cite{cohen2013statistical}. Cohen's $d$ values of 0.2, 0.5, 0.8 generally correspond to small, moderate and large differences between two groups, respectively. A summary of behaviors we found to be significantly different between the two groups can be found in Table \ref{behsummary}. 

In \textit{Setup90}, we found that the maximum duration of exclusions and total duration of exclusions percentage during an interview exhibit significantly different results between the two groups. No significant difference was observed for the \textit{Exclusion} statistics in \textit{Setup45,} supporting the idea that distributing attention was harder in \textit{Setup90} compared to \textit{Setup45} for the ASD participants. Among \textit{Contact} statistics, the average duration of not making contact with anyone was significantly higher in \textit{Setup90} for autistic individuals in comparison with neurotypical individuals. The number of contacts per minute was significantly lower for people with ASD, in both \textit{Setup90} and \textit{Setup45}.

\begin{table}[h]
\centering
\caption{Distribution of head orientations while subject is listening}
\label{listening}
\begin{tabular}{|cl|cc|cc|}
\hline
\multicolumn{2}{|c|}{\multirow{2}{*}{Head Orientation - Listening}} & \multicolumn{2}{c|}{Setup90}         & \multicolumn{2}{c|}{Setup45}         \\ \cline{3-6} 
\multicolumn{2}{|c|}{}                                                  & \multicolumn{1}{c|}{ASD}    & NT     & \multicolumn{1}{c|}{ASD}    & NT     \\ \hline
\multicolumn{2}{|c|}{Interviewer who is speaking}                       & \multicolumn{1}{c|}{27.1\%} & 29.2\% & \multicolumn{1}{c|}{27.4\%} & 28.4\% \\ \hline
\multicolumn{2}{|c|}{Neutral}                                           & \multicolumn{1}{c|}{67.6\%} & 67.4\% & \multicolumn{1}{c|}{65.5\%} & 65.6\% \\ \hline
\multicolumn{2}{|c|}{Other interviewer}                                 & \multicolumn{1}{c|}{4.9\%}  & 3.7\%  & \multicolumn{1}{c|}{5.5\%}  & 4.9\%  \\ \hline
\end{tabular}
\end{table}

\begin{table}[h]
\centering
\caption{Distribution of head orientations while subject is speaking}
\label{speaking}
\begin{tabular}{|cl|cc|cc|}
\hline
\multicolumn{2}{|c|}{\multirow{2}{*}{Head Orientation - Speaking}} & \multicolumn{2}{c|}{Setup90}         & \multicolumn{2}{c|}{Setup45}         \\ \cline{3-6} 
\multicolumn{2}{|c|}{}                                                 & \multicolumn{1}{c|}{ASD}    & NT     & \multicolumn{1}{c|}{ASD}    & NT     \\ \hline
\multicolumn{2}{|c|}{Interviewer who spoke last}                       & \multicolumn{1}{c|}{30.3\%} & 22.2\% & \multicolumn{1}{c|}{22.1\%} & 18.9\% \\ \hline
\multicolumn{2}{|c|}{Neutral}                                          & \multicolumn{1}{c|}{55.2\%} & 60.8\% & \multicolumn{1}{c|}{63.0\%} & 64.7\% \\ \hline
\multicolumn{2}{|c|}{Other interviewer}                                & \multicolumn{1}{c|}{14.1\%} & 17.1\% & \multicolumn{1}{c|}{13.4\%} & 15.3\% \\ \hline
\end{tabular}
\end{table}

In Tables~\ref{listening} and \ref{speaking}, we present the distributions of head orientations based on the subject's conversational role. In this analysis, we again observe that the differences between the two groups are more evident in $Setup90$. The table shows that, when speaking to multiple people, autistic individuals tend to distribute their attention less evenly; their focus usually remains on the person who made the last remark. Neurotypical individuals pay slightly more attention on the person who spoke last, while generally ensuring that the other interviewer is also included in the conversation. This difference is confirmed to be significant by t-tests, as the results reveal that people with ASD tend to look at the interviewer who spoke last significantly more than neurotypical people do. To the best of our knowledge, this is the first quantification of these types of differences about conversational roles and their impact on orienting in triadic settings.

Overall, we conclude that there are noticeable differences between the two groups, and our model is able to reflect and quantify these differences. This is valuable towards the goal of creating a coaching tool which would allow autistic individuals to undertake situational practice. While there are many studies regarding social communication behaviors of autistic people, none of them address these in the context of three-way conversations among adults. The extensive literature on dyadic interactions generally found that autistic individuals, compared with neurotypical individuals,  display behavioral differences such as spending less time looking at other people’s faces \cite{chita2016social} and providing fewer nonverbal cues such as regular eye contact or maintaining a body orientation towards a speaker \cite{bal2019autism}. Based on the literature on dyadic interactions, and the limited literature on triadic interactions for children \cite{mcparland2021investigating, jarrold2013social}, one can expect that autistic adults would display these different attention distribution behaviors in triadic settings as well. As far as we know, we are the first to conduct experiments that characterize differences between neurotypical and autistic adults in triadic conversational settings, and our findings align with these expectations. Although our small and non-random sample does not allow generalization, this preliminary quantification of group differences using an automated system shows an application of our estimation models and is one of our contributions.

\section{Conclusion and Future Work} \label{conclusion}

In this paper, we improve our proposed models in \cite{tepencelik2021body} for body and head orientation estimation that work with low-resolution point clouds generated by two LiDAR sensors. We improve the average error rate of our body orientation estimation model from 8.4 degrees to 5.2 degrees. We enhance our head orientation estimation model by enabling reliable estimations in realistic scenarios where the subject is naturally moving with various head and body poses in a triadic conversation setting. We present novel models that are the first to reliably estimate body and head orientations using LiDAR sensors from surveillance viewpoints. Our estimation results are comparable to results in the literature, although our models work with low-resolution and noisy point clouds and without color information. We also showed that the state-of-the-art models perform poorly in our low-resolution setting although they are effective in high-resolution datasets. This work pushes the boundaries of current body and head orientation estimation systems by demonstrating for the first time that low-resolution, noisy point clouds from LiDAR sensors, without color information, can be used to estimate both body and head orientations from surveillance viewpoints. We believe that accurate orientation estimation models that can work from unobtrusive distances are a significant development. Our proposed models could help with pedestrian safety \cite{rehder2014head, wang2022lidar}, behavior analysis and prediction \cite{chen2012we}, interaction and attention modeling \cite{brvsvcic2013person, liu2007customers, chen2011discovering}, while protecting user privacy.

As an application of our head orientation estimation model, we created a triadic conversation scenario in a room with LiDAR sensors placed to surveillance viewpoints. Using our proposed model, we provide novel analysis on various behaviors in triadic interaction settings and show the differences between autistic and neurotypical individuals using statistical significance tests. We are the first to quantify these qualitatively well-known behavioral differences.

{\bf Limitations of this study;} In future iterations of this work, we can improve data collection; the network overload issue can be prevented by using a sensor that supports a 1 Gbps network instead of 100 Mbps, as well as a more powerful CPU. The estimated head positions from the built-in human detection algorithm will be corrected using our head detection algorithm presented in Section \ref{headpos}. 

An additional limitation of this study is the small size of the dataset; our conversation dataset has 12 autistic and 8 neurotypical individuals. While we found some statistically significant differences, other differences might be revealed with a larger number of subjects. We do not claim that our small and non-random sample is representative of the larger population of employment-seeking autistic young adults. Our purpose with this comparison is to highlight the functionality of the proposed technological contribution that is the orientation estimation models, and it also provides preliminary baseline results for future studies. 

A potential limitation of this study is scalability and robustness to different environments, which we have not experimented with yet for the current study, but plan to explore in our future studies. Another limitation is the environmental control. While all conversations were held in the same room under the same conditions of heat and light and seating arrangements, the naturalistic flow of conversation led to some conversation variability across subjects. Although the interviewers tried to follow conversational scripts closely, minor differences in conversation flow might have affected subject behavior patterns.

{\bf Applications:} The proposed body and head orientation estimation models can be used in various applications. We plan to extend our models to become a component of virtual coaching to high-functioning autistic individuals who are seeking jobs, to integrate them to workplaces. We plan to deploy a behavioral intervention program for autistic individuals using our proposed head orientation estimation and behavior analysis tools so that we can further test their effectiveness, while also getting feedback from our participants about this technology. We also plan to develop automated smart coaches that leverage the differences between neurotypical and autistic individuals, while imitating the decisions of a professional behavioral coach. In conjunction with employer-based initiatives to make workplace environments and hiring practices more autism-friendly, tools that allow situational practice and feedback of social communication could facilitate transition to employment for the large number of autistic individuals aging into adulthood each year.

\vspace{-0.5em}
\section*{Acknowledgment}

We thank Ara Jung, Trent Simmons, Sarah Luo and Saygin Artiran for helping with data collection, Sarah Hacker and Ara Jung for managing the IRB approval and subject payments, and all the participants in our study.

\vspace{-0.5em}
\bibliographystyle{IEEEtran}
\bibliography{IEEEabrv,references}

\begin{IEEEbiography}[{\includegraphics[width=1in,height=1.333in,clip,keepaspectratio]{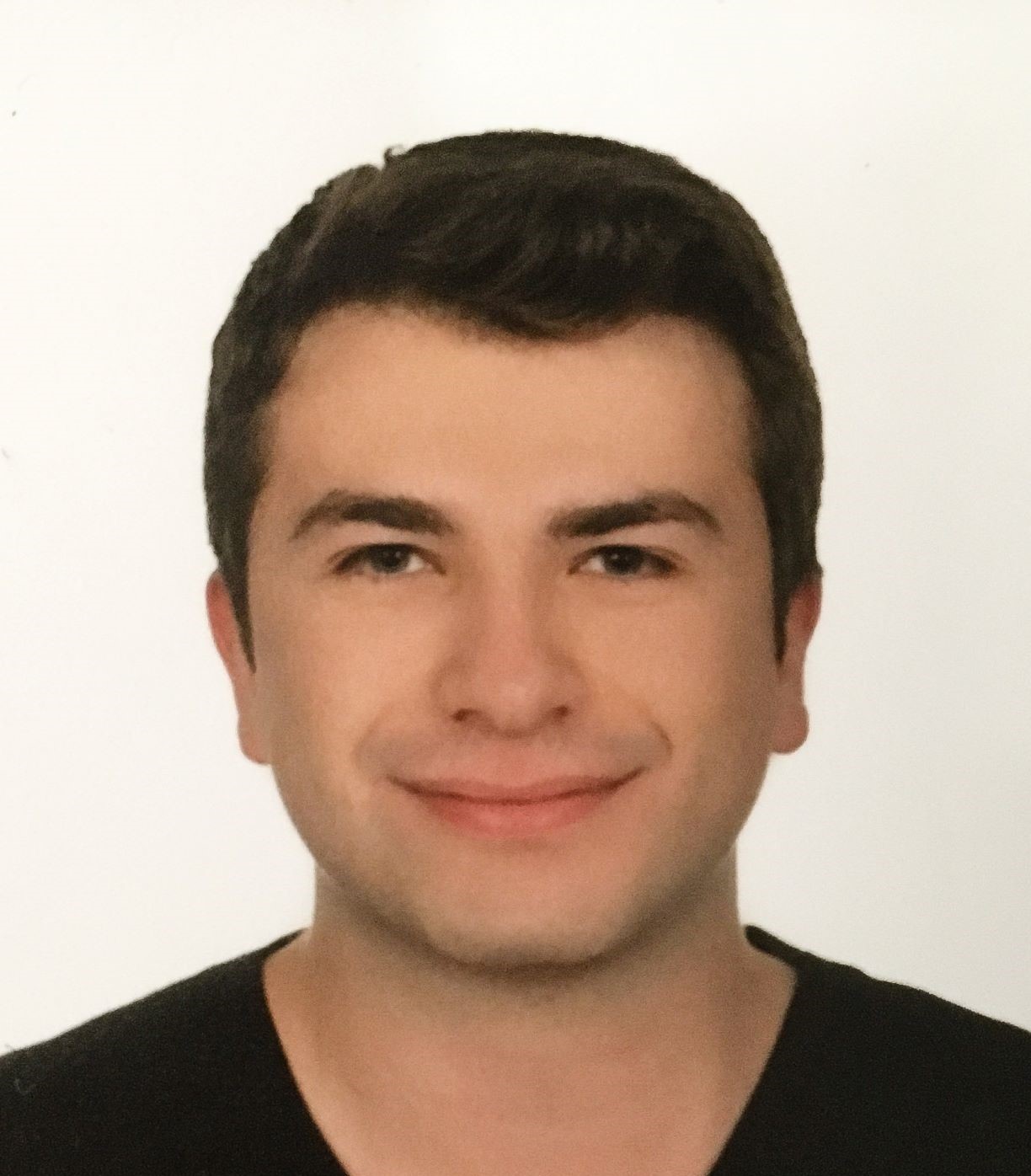}}]{Onur Tepencelik} received the B.S. degree in electrical and electronics engineering from Bilkent Unibersity, Turkey, in 2019. He received the M.S. degreee in electrical and computer engineering from University of California, San Diego, in 2021, where he is pursuing the Ph.D. degree.

He has been a Graduate Student Researcher at the Mobile Systems Design Laboratory, University of California, San Diego, since 2020. His research interests include machine learning, computer vision, 3D processing and LiDAR sensors with a focus in human head and body pose estimation. 
\end{IEEEbiography}

\begin{IEEEbiography}[{\includegraphics[width=1in,height=1.333in,clip,keepaspectratio]{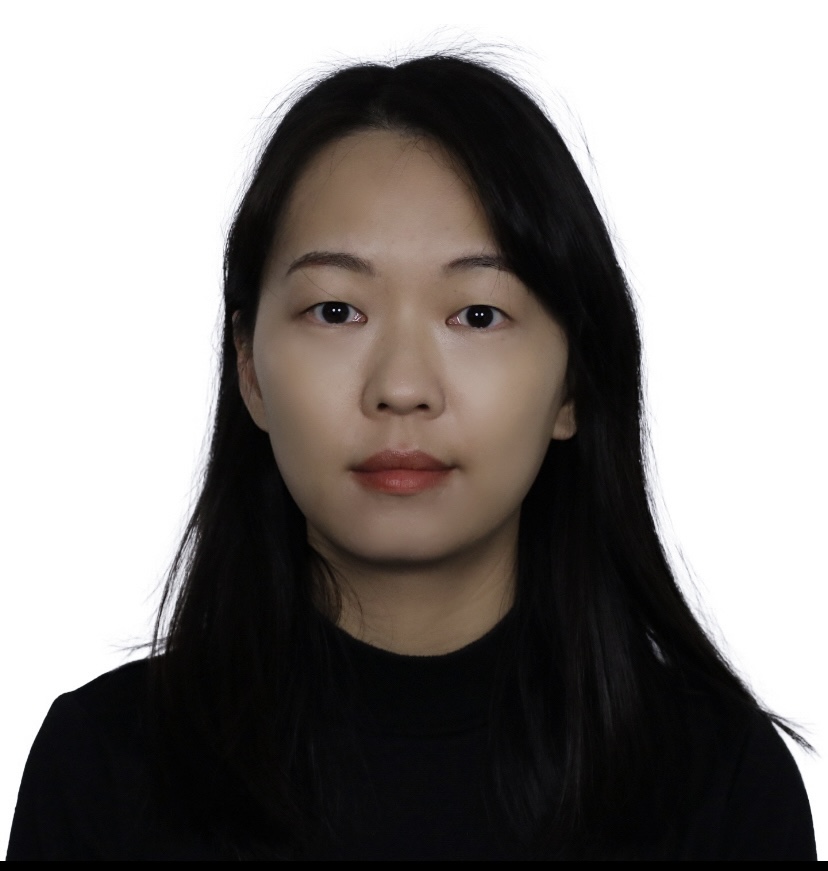}}]{Wenchuan Wei} received the B.S. degree in electronic cngineering from Tsinghua University in 2014 and Ph.D. degree in computer engineering from University of California, San Diego, in 2020. 

She is currently working as an AR/VR software engineer at Apple Inc., USA. Her research interests include AR/VR, machine learning, digital health, and multimedia.
\end{IEEEbiography}

\begin{IEEEbiography}[{\includegraphics[width=1in,height=1.333in,clip,keepaspectratio]{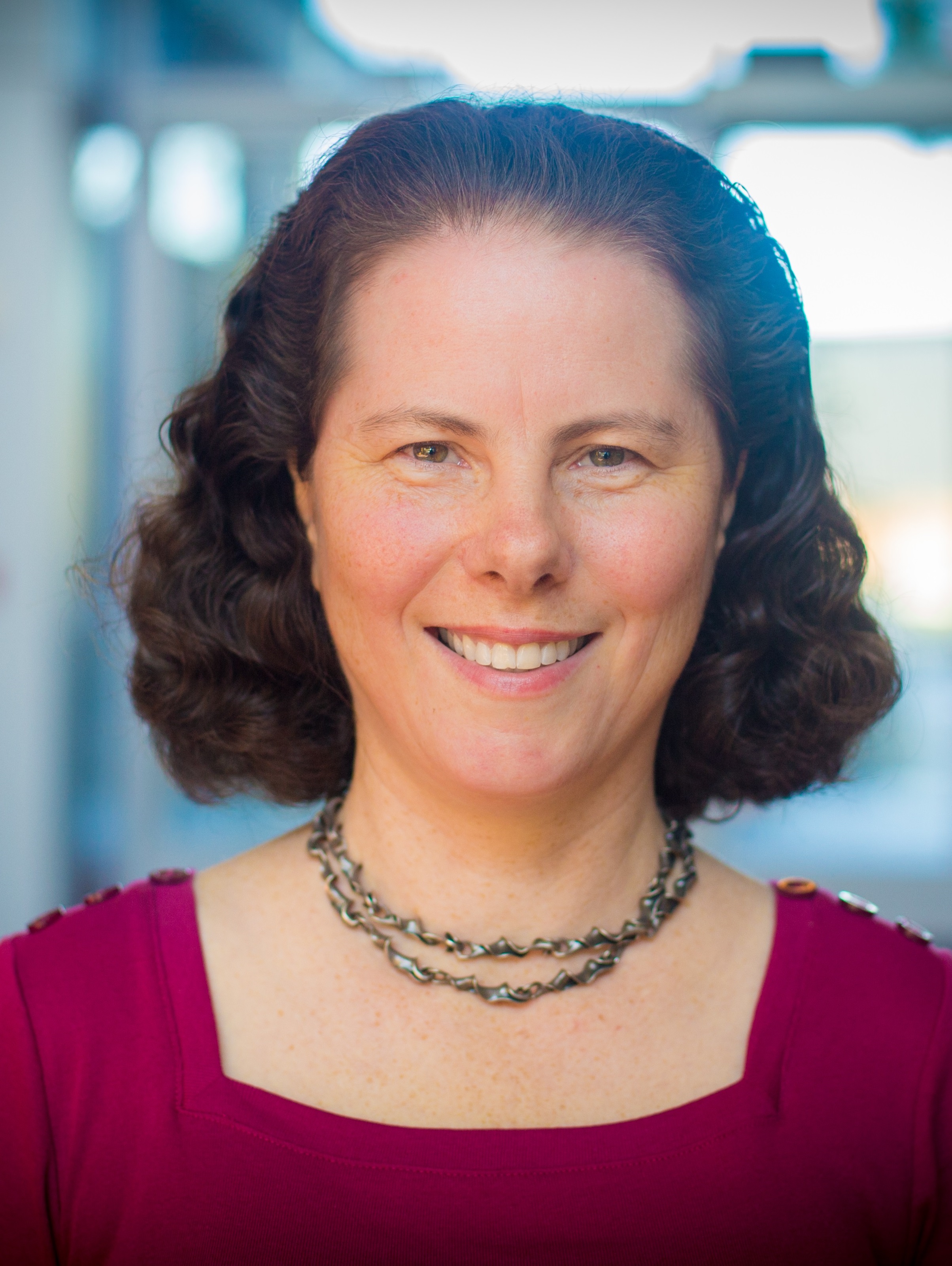}}]{Pamela Cosman} (Fellow, IEEE) received the B.S. degree with Honor in electrical engineering from the California Institute of Technology in 1987, and the Ph.D. degree in electrical engineering from Stanford University in 1993.  Following an NSF Postdoctoral Fellowship at Stanford and the University of Minnesota from 1993 to 1995, she joined the Faculty of the Department of Electrical and Computer Engineering at UC San Diego, where she is currently a Distinguished Professor. Her administrative positions include serving as the Director of the Center for Wireless Communications (2006-08), ECE Department vice chair (2011-14), and Associate Dean for Students (2013-16). She has written over 300 technical articles in the areas of image and video compression and processing, and wireless communications.  She is also the author of two children’s books, The Secret Code Menace and The Hexagon Clue, that introduce error correction coding and other technical concepts through fiction. 

Dr. Cosman is a member of Tau Beta Pi and Sigma Xi. She has been a member of the Technical Program Committee or the Organizing Committee for numerous conferences, including serving as Technical Program Co-Chair of ICME 2018. Her awards include the Electrical and Computer Engineering Departmental Graduate Teaching Award, the Career Award from the National Science Foundation, and the 2018 National Diversity Award from the Electrical and Computer Engineering Department Heads Association. She was an Associate Editor of the IEEE Communications Letters and of the IEEE Signal Processing Letters.  She was the Editor-in-Chief (2006-09) and a Senior Editor (2003-05 and 2010-13) of the IEEE Journal on Selected Areas in Communications.
\end{IEEEbiography}

\begin{IEEEbiography}[{\includegraphics[width=1in,height=1.333in,clip,keepaspectratio]{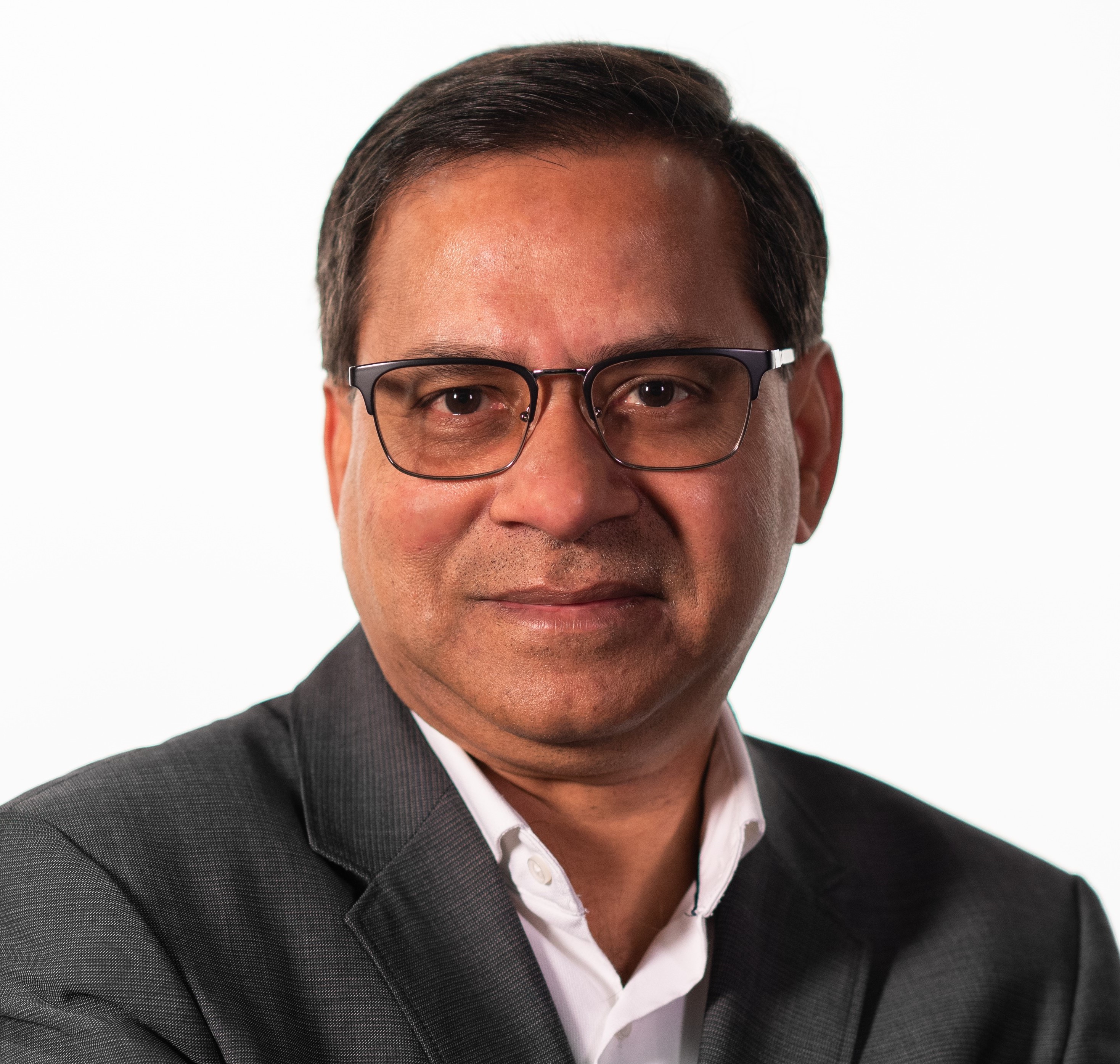}}]{Sujit Dey} (Fellow, IEEE) received the Ph.D. degree in computer science from Duke University, in 1991. 

In 2004, he founded Ortiva Wireless, where he has served as its founding CEO and later as the CTO and Chief Technologist till its acquisition by Allot Communications, in 2012. Prior to Ortiva, he served as the Chair of the Advisory Board of Zyray Wireless till its acquisition by Broadcom, in 2004, and an advisor to multiple companies, including ST Microelectronics and NEC. He has served as the Faculty Director of the von Liebig Entrepreneurism Center, from 2013 to 2015, and the Chief Scientist, Mobile Networks, at Allot Communications, from 2012 to 2013. In 2015, he co-founded igrenEnergi Inc., providing intelligent battery technology and solutions for EV mobility services. He heads the Mobile Systems Design Laboratory, developing innovative and sustainable edge computing, networking and communications, multi-modal sensor fusion, and deep learning algorithms and architectures to enable predictive personalized health, immersive multimedia, and smart transportation applications. He has created inter-disciplinary programs involving multiple UCSD schools as well as community, city, and industry partners; notably the Connected Health Program, in 2016, and the Smart Transportation Innovation Program, in 2018. Prior to joining UCSD in 1997, he was a Senior Research Staff Member at NEC C\&C Research Laboratories, Princeton, NJ, USA. In 2017, he was appointed as an Adjunct Professor at the Rady School of Management and the Jacobs Family Endowed Chair in Engineering Management Leadership. He is currently a Professor with the Department of Electrical and Computer Engineering and the Director of the Center for Wireless Communications and the Institute for the Global Entrepreneur, University of California, San Diego. He has coauthored more than 250 publications, and a book on Low-Power Design. He holds 18 U.S. and two international patents, resulting in multiple technology licensing and commercialization. 

Dr. Dey has been a recipient of nine IEEE/ACM Best Paper Awards, and has chaired multiple IEEE conferences and workshops.
\end{IEEEbiography}

\end{document}